\documentclass[letterpaper]{article} % DO NOT CHANGE THIS
\usepackage{aaai2027}  % DO NOT CHANGE THIS
\usepackage[hyphens]{url}  % DO NOT CHANGE THIS
\usepackage{graphicx} % DO NOT CHANGE THIS
\usepackage{natbib}  % DO NOT CHANGE THIS AND DO NOT ADD ANY OPTIONS TO IT
\usepackage{caption} % DO NOT CHANGE THIS AND DO NOT ADD ANY OPTIONS TO IT
\usepackage{algorithm}
\usepackage{algorithmic}

\usepackage{newfloat}
\usepackage{listings}
\DeclareCaptionStyle{ruled}{labelfont=normalfont,labelsep=colon,strut=off} % DO NOT CHANGE THIS
\floatstyle{ruled}
\newfloat{listing}{tb}{lst}{}
\floatname{listing}{Listing}

\usepackage{booktabs}
\usepackage{amsmath}
\usepackage{amssymb}
\usepackage[table]{xcolor}
\usepackage{multirow}

\title{OmniMate: Open-Ended Real-Time Streaming Audio-Visual Generation for Interactive Avatars}
\author{
    Quanyue Song\textsuperscript{\rm 1,\rm 2}\thanks{Work done during an internship at China Telecom Artificial Intelligence Technology (Beijing) Co., Ltd.}, 
    Yishan He\textsuperscript{\rm 2},
    Yanbo Ding\textsuperscript{\rm 2,\rm 3},\\
    Zhixiang He\textsuperscript{\rm 2},
    Yongxiang Li\textsuperscript{\rm 2},
    Caigui Jiang\textsuperscript{\rm 1}\corresponding,
    Zhizhi Guo\textsuperscript{\rm 2}\corresponding
}

\affiliations{
    \textsuperscript{\rm 1}State Key Laboratory of Human-Machine Hybrid Augmented Intelligence, Institute of Artificial Intelligence and Robotics, Xi'an Jiaotong University, China\\
    \textsuperscript{\rm 2}China Telecom Artificial Intelligence Technology (Beijing) Co., Ltd., China\\
    \textsuperscript{\rm 3}Shenzhen Key Laboratory of Computer Vision and Pattern Recognition, Shenzhen Institutes
of Advanced Technology, Chinese Academy of Sciences, Shenzhen, China
}

\begin{document}

\maketitle

\begin{abstract}
Recent advances in diffusion-based generative models have enabled real-time audio-driven avatar generation and unified audio-visual synthesis, providing a promising foundation for interactive avatar systems. However, extending unified audio-visual synthesis to real-time interactive streaming remains challenging, as the generation horizon is unknown in advance and the generated identity may drift over long-term generation. To address these challenges, we propose OmniMate, a unified framework for open-ended real-time interactive audio-visual avatar generation. OmniMate jointly synthesizes visual content, speech, and sound effects in real time, enabling natural and immersive multi-turn interactions. To achieve adaptive response progression, we introduce a Generation Progress Controller (GPC) that explicitly models the generation progress of each streaming chunk, allowing the model to complete responses according to the desired progress and achieve seamless transitions between execution and listening states. To preserve long-term cross-modal identity consistency, we propose a Multi-Reference Conditioning Module (MRCM), which leverages multiple reference images and a reference speech segment to provide persistent visual and speaker identity cues throughout long-duration streaming interactions. Extensive experiments on an interaction-oriented adaptation of VerseBench demonstrate that OmniMate achieves high-quality, low-latency streaming generation while maintaining strong long-term audio-visual consistency. The results further show that OmniMate supports realistic, coherent, and responsive interactive avatar experiences over extended multi-turn conversations.
\end{abstract}


\begin{figure*}[t]
\centering
\includegraphics[width=2\columnwidth, trim=20 290 20 30, clip=true]{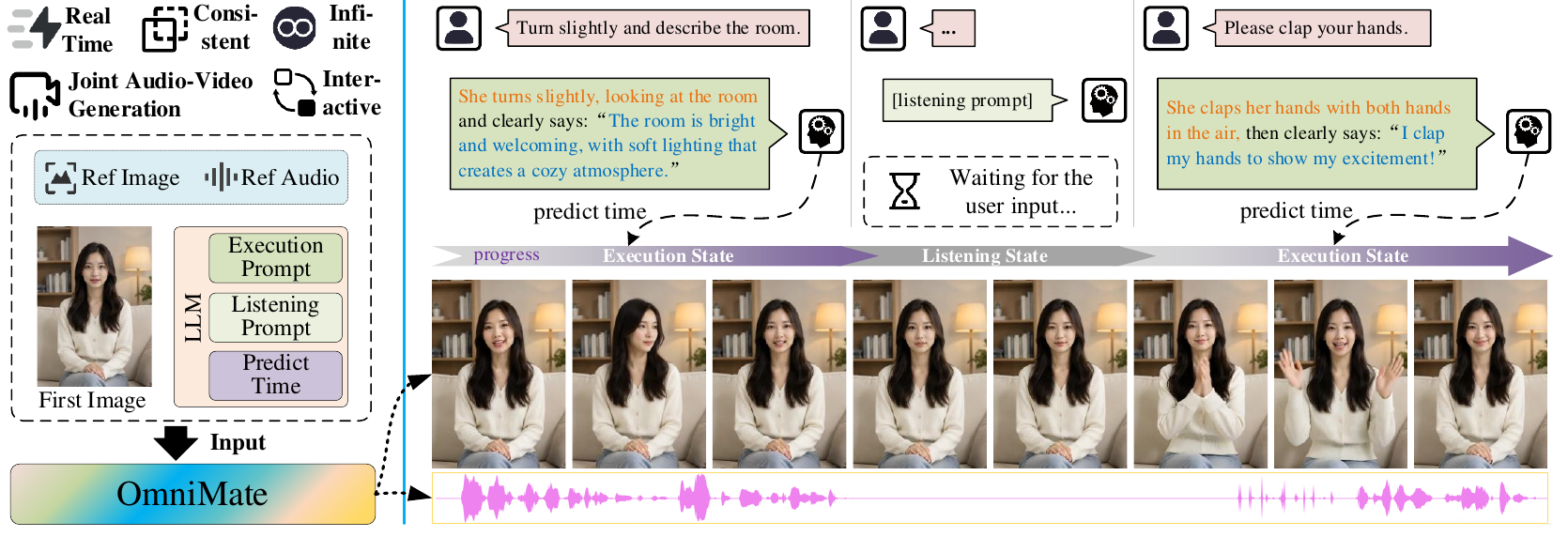} 
\caption{We propose OmniMate, an open-ended real-time streaming audio-visual interactive avatar generation framework, which jointly generates speech, motions, and sound effects while preserving visual and speaker identity across multi-turn conversations.}
\label{fig1}
\end{figure*}

\section{Introduction}
Recent advances in diffusion-based generative models have enabled high-fidelity human-centric video synthesis~\cite{wan2025wan}. Meanwhile, diffusion distillation techniques~\cite{yin2025slow} have brought audio-driven video generation towards real-time inference, facilitating interactive avatar applications~\cite{huang2025live}. More recently, unified audio-visual models~\cite{hacohen2026ltx,team2026mova} have enabled joint synthesis of synchronized speech, sounds, and visual content. However, extending such joint generation to open-ended real-time streaming remains non-trivial due to interaction requirements and identity preservation.

The first challenge lies in determining the generation horizon during open-ended interactive streaming. Unlike conventional audio-driven video generation with predefined speech segments, interactive audio-visual streaming requires the model to continuously generate responses based on execution prompts without explicit knowledge of whether the instructed content has been fully expressed. As a result, the model may fail to determine when to terminate the current execution state and transition back to the listening state. A fixed generation duration may truncate incomplete responses or produce redundant speech and motions, while explicit completion detection introduces additional latency, limiting real-time interaction.

The second challenge concerns long-term cross-modal identity consistency in streaming generation. Due to limited temporal context, identity cues gradually fade over time. Single-image or short temporal adjacent chunks struggle to preserve visual identity under unseen poses and viewpoints, leading to identity drift during long-term generation. Meanwhile, limited speech context makes it difficult to maintain consistent voice timbre across multiple conversational turns.

To address these challenges, we propose OmniMate, a unified framework for open-ended, real-time interactive audio-visual avatar generation (see Figure~\ref{fig1}). Given user instructions, OmniMate jointly generates speech, facial motions, body gestures, and action-related sound effects (e.g., applause) conditioned on user instructions, enabling immersive interactions. It further preserves long-term visual identity and speaker consistency throughout open-ended streaming generation. Specifically, OmniMate introduces a Generation Progress Controller (GPC) for adaptive generation termination and a Multi-Reference Conditioning Module (MRCM) for cross-modal identity preservation within a unified streaming framework.

The GPC enables seamless transitions between execution and listening states during multi-turn streaming interactions. By providing each generated chunk with an explicit notion of generation progress, GPC allows the model to perceive both the overall conversational intent and its relative progress within the current response, enabling duration-aware response generation according to the desired schedule and seamless transition back to the listening state. Furthermore, the progress-aware conditioning suppresses the influence of outdated historical chunks from previous state turns, enabling immediate execution of newly arrived user instructions and low-latency state transitions.

The MRCM preserves long-term cross-modal identity consistency by leveraging multiple reference images and a reference speech segment. Compared with a single reference image, multiple views provide richer identity cues, improving visual identity preservation under large pose and viewpoint variations during streaming generation. Meanwhile, the reference speech segment captures the target speaker’s vocal characteristics, enabling consistent speaker identity and stable voice timbre across multi-turn interactions.

By integrating these two components into a unified streaming generation framework, OmniMate enables real-time synthesis of interactive audio-visual avatars with strong cross-modal consistency. We evaluate OmniMate on an interaction-oriented version of the VerseBench~\cite{wang2025universe1unifiedaudiovideogeneration}, assessing its generation quality, responsiveness, and long-term consistency. Both qualitative and quantitative results demonstrate that OmniMate achieves low-latency, real-time streaming audio-visual generation while maintaining long-term visual identity and speaker consistency over extended interactions, and effectively supports realistic and user-avatar conversational behaviors with competitive performance.

Contributions of our OmniMate are listed below:

\begin{itemize}
\item We propose OmniMate, an open-ended real-time interactive audio-visual avatar generation framework, capable of jointly synthesizing speech, facial expressions, body gestures, and environmental sounds while maintaining low-latency streaming interactions.
\item We introduce a Generation Progress Controller (GPC) that explicitly models the generation progress of each streaming chunk, enabling responses to follow the desired execution schedule, seamless transitions between execution and listening states, and prompt responses to newly arrived user instructions with minimal latency.
\item We design a Multi-Reference Conditioning Module (MRCM) that leverages multiple reference images and a reference speech segment to preserve long-term visual identity and speaker consistency throughout open-ended streaming generation.
\end{itemize}

\section{Related Work}

\begin{figure*}[t]
\centering
\includegraphics[width=2\columnwidth, trim=20 250 20 30, clip=true]{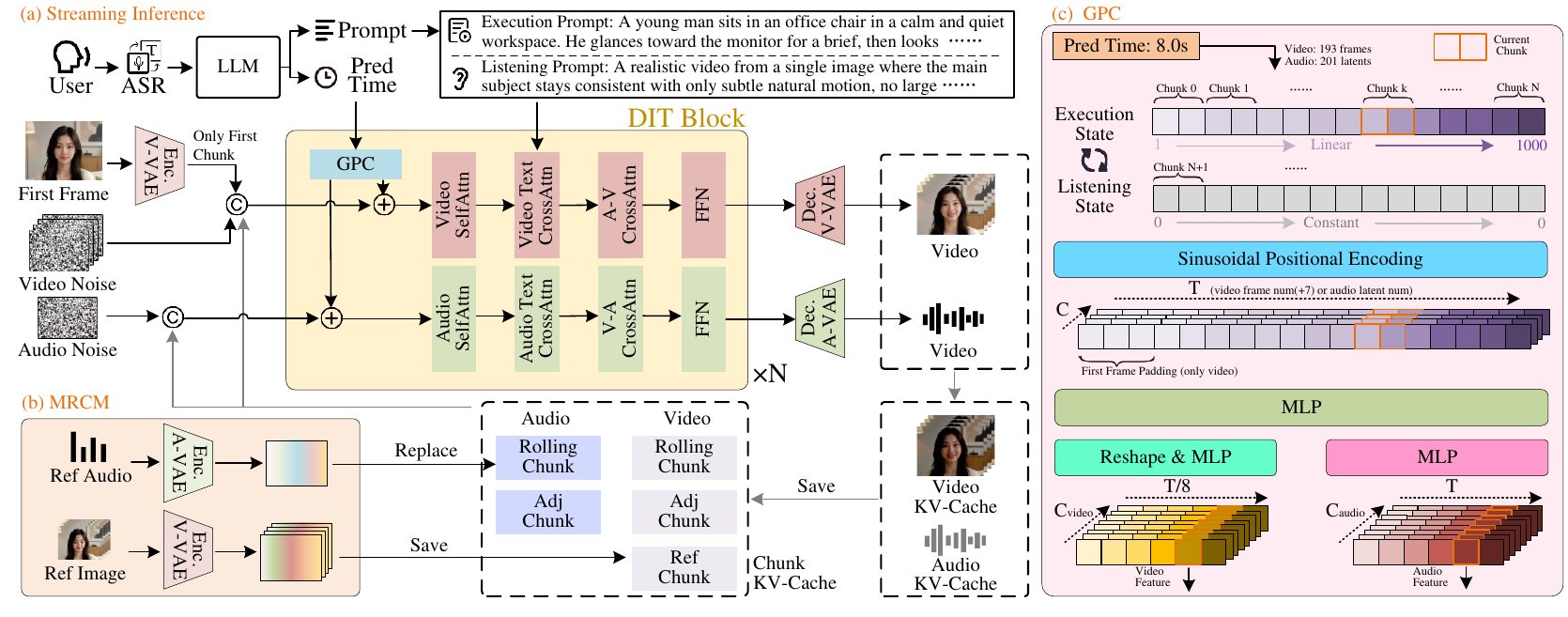} 
\caption{Overview of OmniMate. (a) Streaming inference pipeline of our framework for real-time interactive audio-visual generation. (b) The Generation Progress Controller (GPC) enables generation progress awareness and seamless state transitions during streaming generation. (c) The Multi-Reference Conditioning Module (MRCM) preserves visual identity and speaker consistency during long-term generation.}
\label{fig2}
\end{figure*}

\subsection{Joint Audio-Video Generation}
Recent advances have extended generative models from audio-to-video synthesis~\cite{wang2025fantasytalking,kong2025let} to joint audio-video generation. A line of works~\cite{cheng2026unisonharmonizingmotionspeech,zhang2026uniavgen,low2025ovi,hacohen2026ltx,team2026mova} adopts separate video and audio branches with cross-attention mechanisms to enable cross-modal interaction. Another line of research~\cite{huang2025jova,chern2026speed} explores unified architectures that jointly model audio and video within a single feature space. Benefiting from large-scale audio-visual training data, these models demonstrate strong capabilities in producing synchronized and realistic audio-visual outputs. Among existing foundation models, LTX2.3~\cite{hacohen2026ltx} achieves a favorable balance between generation quality and efficiency, and therefore serves as the backbone of our framework.

\subsection{Streaming Video Generation}
To enable streaming inference, several works~\cite{huang2025live,xie2025x,sun2025streamavatar,yang2025longlive} distill bidirectional diffusion models into causal architectures~\cite{yin2025slow}, leveraging techniques such as distribution matching distillation~\cite{Yin_2024_CVPR,yin2024improved} and self-forcing training~\cite{huang2025self,cui2025self}. Meanwhile, recent studies~\cite{wang2026flowact,ng2022learning,wang2025diffusion} introduce listening states or predefined actions, and advanced approaches~\cite{song2026interactiveavatar} have extended streaming generation to interactive avatar scenarios, enabling speech-driven responses with synchronized facial and body motions. More recently, StreamChar~\cite{tian2026streamchar} and WanStream~\cite{huang2026wan} demonstrate real-time interactive audio-video generation, however, they rely on large-scale training data or require training additional models with a substantial number of parameters. In contrast, we extend streaming generation to real-time joint audio-video synthesis for immersive interactive experiences through a lightweight adaptation.

\section{Method}

\begin{figure*}[t]
\centering
\includegraphics[width=2\columnwidth, trim=40 300 40 20, clip=true]{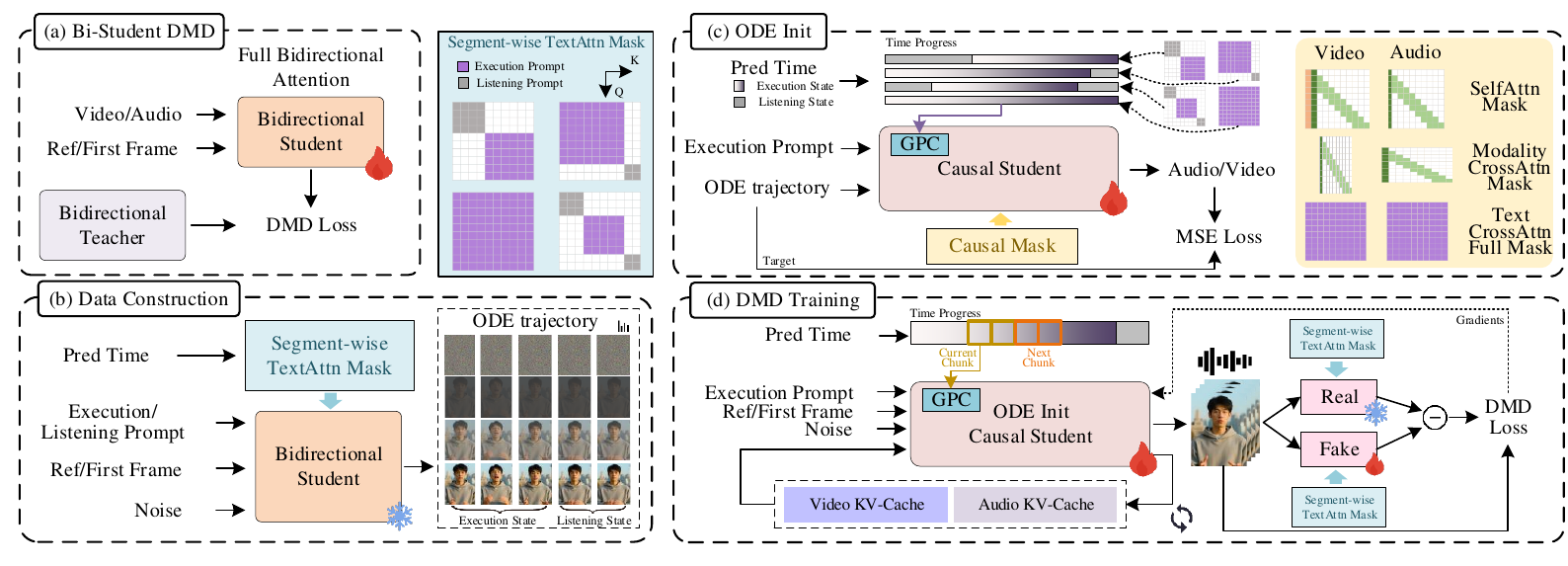} 
\caption{Training procedure of GPC. (a) A DMD-distilled bidirectional teacher model is used to (b) generate ODE trajectories with state transitions. (c) ODE initialization provides a better starting point for the causal student model, and (d) DMD further distills the causal student model for real-time streaming generation.}
\label{fig3}
\end{figure*}

\subsection{Overview}
As illustrated in Figure~\ref{fig2}, given a first frame, optional reference images and audio, OmniMate generates real-time streaming audio-visual content in response to user instructions while supporting seamless switching between listening and execution states. User speech is transcribed by an ASR system and processed by the LLM to produce response content, action instructions, and the estimated execution duration. The duration is provided to the Generation Progress Controller (GPC) to guide response streaming generation. Meanwhile, the Multi-Reference Conditioning Module (MRCM) integrates visual and audio identity cues to preserve character appearance and speaker consistency during long-term generation. Combined with DMD-based distillation, OmniMate enables open-ended, identity-consistent, and interactive real-time audio-visual generation.

\subsection{Generation Progress Controller}
To support complete and coherent streaming generation under limited historical context, we propose the Generation Progress Controller (GPC), which provides explicit awareness of the generation progress within the current instruction. GPC maps latent-aligned progress into a continuous time embedding and injects it into the model by adding it to the latent representations, guiding chunk-wise generation and preventing incomplete execution or redundant generation. During training, we leverage videos generated by a distilled bidirectional teacher model with execution and listening state transitions, and supervise the model to learn progress-aware temporal representations.

\subsubsection{Generation Progress Conditioning}
To enable the model to perceive the execution progress of each instruction during streaming generation, GPC encodes temporal progress as continuous conditioning signals. Given the LLM-estimated execution duration, we first determine the required number of generated tokens and assign progress values ranging from 1 to 1000. For video frames, the progress value is defined as:
\begin{equation}
p_i^v = 1+\frac{i-1}{T_v-1}\times999,\quad i=1,\dots,T_v
\end{equation}
where $T_v$ denotes the number of video frames.

Since the first video latent corresponds to a single frame while subsequent latents represent 8 frames, we align frame-level progress with video latents by repeating the first-frame progress value eight times and grouping every eight frame-level embeddings into one latent token. Specifically, the frame-level progress embedding is transformed as $[B,T_v+7,D]\rightarrow[B,(T_v+7)/8,8D]$, where the first latent token contains the repeated progress embedding of the first frame and the remaining tokens aggregate embeddings from consecutive 8-frame intervals.

Similarly, the audio progress is defined over audio tokens:
\begin{equation}
p_j^a = 1+\frac{j-1}{T_a-1}\times999,\quad j=1,\dots,T_a
\end{equation}
where $T_a$ denotes the number of audio tokens. 

Tokens beyond the predicted execution duration are assigned a progress value of 0, indicating the listening state.

Inspired by OmniAvatar~\cite{gan2025omniavatar}, the progress signals of both modalities are encoded by Sinusoidal Positional Encoding~\cite{vaswani2017attention} and a multi-layer MLP, and then injected into the corresponding latents:
\begin{equation}
\tilde{z}^{m}=z^{m}+\mathrm{MLP}(\mathrm{PE}(p^{m})),\quad m\in\{a,v\}
\end{equation}
where $z^m$ and $\tilde{z}^m$ denote the original and progress-conditioned latents, respectively. By providing explicit awareness of execution progress, GPC enables the model to complete responses according to the desired schedule and seamlessly transition between execution and listening states.

\subsubsection{State-Transition Data Construction}
To train GPC with explicit execution-listening transitions, we construct state-transition training data using the DMD-distilled bidirectional teacher during ODE initialization. Specifically, the teacher generates sequences containing both execution and listening states, while latent trajectories at different denoising timesteps are recorded as supervision (Figure~\ref{fig3}.a and~\ref{fig3}.b).

Given a transition boundary $\tau$, we define a binary segment-wise text cross attention mask as illustrated in Figure~\ref{fig3}:
\begin{equation}
m_i=
\begin{cases}
1, & i < \tau,\\
0, & \text{otherwise},
\end{cases}
\end{equation}
where $m_i=1$ denotes the execution state and $m_i=0$ denotes the listening state. By varying the initial state and transition boundaries, we construct diverse state-transition sequences, including only execution, execution-to-listening, listening-to-execution, and listening-to-execution-to-listening transitions.

For audio-text and video-text cross-attention, execution-state latents attend to the execution prompt $P_{\mathrm{exec}}$, while listening-state latents attend to a generic listening prompt $P_{\mathrm{listen}}$. The condition associated with the $i$-th latent is defined as:
\begin{equation}
P_i=m_iP_{\mathrm{exec}}+(1-m_i)P_{\mathrm{listen}}
\end{equation}
This enables the teacher to generate temporally continuous sequences with clear state transitions, providing reliable supervision for progress learning.

We adopt the DMD-distilled bidirectional teacher instead of the original multi-step diffusion teacher, as the latter suffers from temporal prompt leakage. Specifically, semantic information may propagate across the transition boundary, causing the actual state transition to deviate from the predefined boundary and resulting in inaccurate supervision.

\subsubsection{Joint Training with DMD}
During the ODE initialization stage, we jointly optimize GPC and the student model. Specifically, the student’s video-text and audio-text cross-attention layers are conditioned only on the execution prompt for the entire sequence, while state transitions are controlled exclusively by the injected generation progress (Figure~\ref{fig3}.c). The progress values are assigned from 1 to 1000 during execution and set to 0 during listening states. Supervised by the state-transition sequences constructed in the previous section, the model learns to perform execution-listening transitions based on generation progress while maintaining coherent streaming generation.

During the subsequent DMD distillation stage, the student retains the same progress conditioning strategy learned during ODE initialization(Figure~\ref{fig3}.d). Meanwhile, the real and fake score networks adopt state-aware text conditioning, where cross-attention layers attend to execution or listening prompts according to the corresponding temporal tokens. This design allows DMD to transfer the teacher’s state-transition behavior to the student, while enabling the student to control state transitions solely through generation progress conditioning.

\subsection{Multi-Reference Conditioning Module}
To preserve long-term cross-modal identity consistency during streaming generation, we propose the Multi-Reference Conditioning Module (MRCM), which provides persistent visual and audio identity cues. MRCM leverages multiple reference images and a reference audio to maintain consistent avatar appearance and speaker characteristics throughout long-term audio-visual generation.

\subsubsection{Visual Identity Conditioning}
We introduce a multi-reference visual conditioning strategy to mitigate visual identity drift. Given $N$ reference images $\{I_i\}_{i=1}^{N}$, we first encode them into latent representations:
\begin{equation}
z_i^{ref}=E_{\mathrm{VAE}}(I_i), \quad i=1,\dots,N
\end{equation}
The reference latents are concatenated with noisy video latents along the temporal dimension:
\begin{equation}
z_t^{in}=[z_1^{ref},z_2^{ref},...,z_N^{ref},z_t]
\end{equation}
where $z_t$ denotes the noisy latent of the generated video. To avoid affecting the temporal positions of generated frames, negative RoPE~\cite{su2024roformer} is applied to reference tokens. During training, four identity-related frames are randomly sampled as references. During inference, up to four reference images are used, with duplication applied when fewer references are available.

\subsubsection{Speaker Identity Conditioning}
We further introduce a training-free speaker identity conditioning strategy. Specifically, we initialize the generation process with a fixed warm-up utterance produced by the model. Given an optional reference audio, the corresponding audio latents are replaced with reference speech latents during denoising:
\begin{equation}
z_t^{a} =
\begin{cases}
z_{\mathrm{ref}}^{a}, & \text{with reference audio},\\
z_{\mathrm{gen}}^{a}, & \text{otherwise},
\end{cases}
\end{equation}
where $z_{\mathrm{ref}}^{a}$ and $z_{\mathrm{gen}}^{a}$ denote reference and generated audio latents. In the absence of a reference audio, the original generated audio is retained without additional constraints. By continuously preserving reference audio latents during streaming generation, MRCM maintains stable speaker characteristics across long audio-visual sequences.

\subsection{Training and Streaming Inference}
The training process consists of four stages. We first fine-tune LTX2.3 with visual identity conditioning. Next, we distill a few-step bidirectional teacher model using DMD and construct state-transition ODE trajectories for initialization. We then perform ODE initialization to jointly optimize GPC and the student model with causal attention mask. Finally, we further distill the initialized model using self-forcing DMD. Specifically, each generated chunk is only allowed to attend to the reference chunk, the first chunk, and two adjacent chunks, enforcing autoregressive generation behavior while improving long-term temporal continuity and stability.

During streaming inference, we use LLM to generate response content and estimated execution duration based on user input. Then, our trained model takes a first frame as input, while optional reference images and audio are processed by MRCM for persistent identity conditioning. The estimated duration is converted into progress signals by GPC, which guides streaming generation and enables transitions between execution and listening states. Meanwhile, we use RollingSink~\cite{li2026rolling} to maintain visual stability during long-term streaming generation.

\section{Experiment}

\begin{table*}[!t]
\centering
\resizebox{\textwidth}{!}{
\begin{tabular}{lccccccccccccc}
\toprule
\multirow{2}{*}{\textbf{Model}} & \multicolumn{2}{c}{\textbf{Speed}} & \multicolumn{4}{c}{\textbf{Audio}} & \multicolumn{3}{c}{\textbf{Video}} & \multicolumn{2}{c}{\textbf{Audio-Video}} & \multicolumn{2}{c}{\textbf{Lip-Sync}} \\
\cmidrule(lr){2-3} \cmidrule(lr){4-7} \cmidrule(lr){8-10} \cmidrule(lr){11-12} \cmidrule(lr){13-14}
 & FPS\(\uparrow\) &TTFF\(\downarrow\) & AQ\(\uparrow\) & A-ID\(\uparrow\)& Speech\(\downarrow\) & TA\(\uparrow\) & VQ\(\uparrow\)& TV\(\uparrow\) & V-ID\(\uparrow\) & Sync-D\(\downarrow\) & IB-Score\(\uparrow\) & LSE-C\(\uparrow\) & LSE-D\(\downarrow\)\\
\midrule
Ovi-1.1 & 2.14 & 114s & 5.65 & 0.353 & 0.306 & \underline{0.543} & 3.93 & 25.27 & 0.932 & 0.112 & 0.200 & \textbf{4.12} & \underline{7.49} \\
MOVA-360P & 0.29 & 658s & \textbf{6.46} & 0.327 & 0.194 & \textbf{0.557} & \textbf{3.97} & 25.26 & 0.937 & \underline{0.018} & \underline{0.345} & 3.59 & 8.06 \\
UniVerse-1 & 0.82 & 143s & 4.77 & 0.471 & 0.271 & 0.347 & 3.61 & 25.17 & \underline{0.953} & 0.250 & 0.192 & 2.55 & 9.14 \\
DaVinci-Base & 4.27 & 45s & 5.95 & 0.339 & 0.277 & 0.475 & 3.68 & \textbf{25.43} & 0.931 & 0.034 & 0.226 & 3.17 & 7.58 \\
LTX-2.3 & 2.59 & 76s & \underline{6.39} & 0.464 & \underline{0.121} & 0.401 & 3.69 & \underline{25.30} & 0.941 & 0.028 & 0.245 & 3.81 & \textbf{6.44} \\
\arrayrulecolor{gray!70}
\midrule
\arrayrulecolor{black}
OmniForcing & \underline{24.83} & \underline{3.51s} & 5.92 & 0.487 & 0.675 & 0.246 & 3.85 & 24.32 & 0.952 & 0.922 & 0.208 & 2.85 & 9.67 \\
Hallo-Live & 19.22 & 3.63s & 5.43 & \textbf{0.593} & 0.773 & 0.443 & 3.77 & 25.30 & 0.950 & \textbf{0.004} & 0.192 & 1.51 & 7.66 \\
\arrayrulecolor{gray!70}
\midrule
\arrayrulecolor{black}
\textbf{OmniMate(Ours)} & \textbf{27.64} & \textbf{3.49s} & 6.22 & \underline{0.529} & \textbf{0.097} & 0.449 & \underline{3.93} & 25.19 & \textbf{0.957} & 0.042 & \textbf{0.356} & \underline{4.06} & 8.66 \\
\bottomrule
\end{tabular}
}
\caption{Quantitative comparison with state-of-the-art methods, including real-time and non-real-time approaches. Best in \textbf{bold} and second best \underline{underlined}. Experiments are conducted using the H100 GPU.}
\label{tab:result}
\end{table*}

\begin{figure*}[t]
\centering
\includegraphics[width=2\columnwidth, trim=15 270 15 30, clip=true]{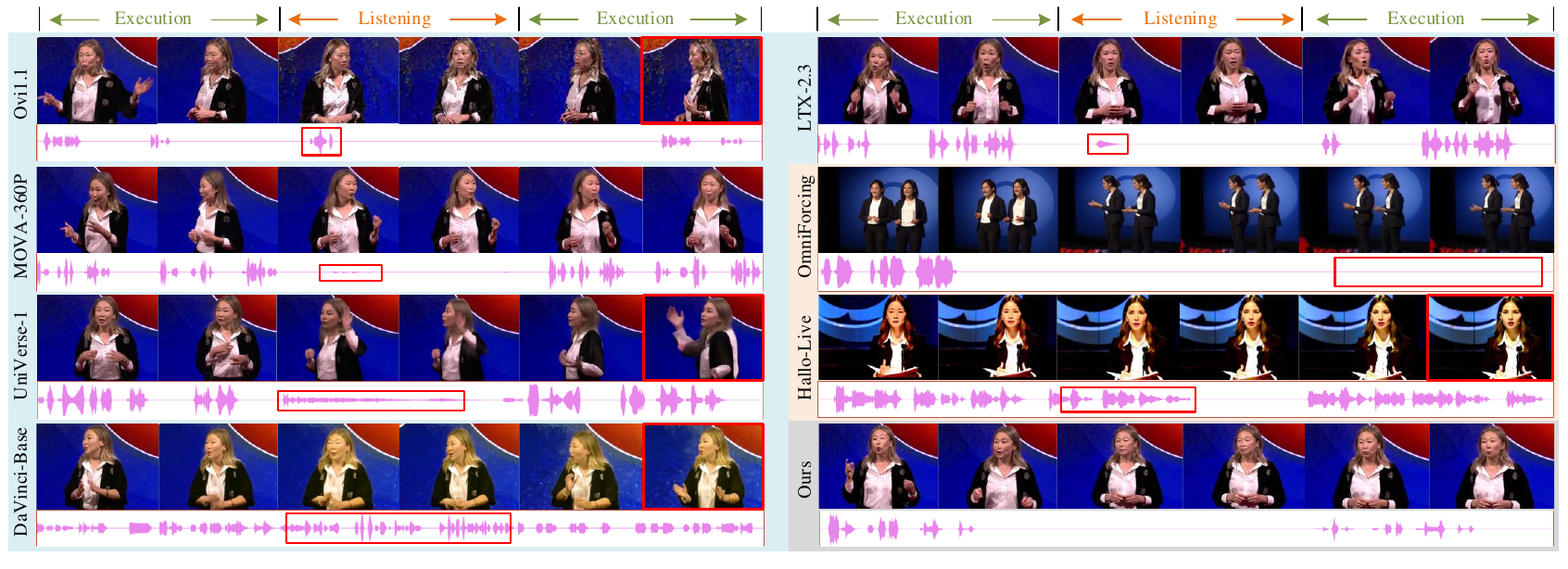} 
\caption{Qualitative comparisons with state-of-the-art methods. Our method demonstrates smooth transitions between interaction states while achieving high-fidelity audio-visual generation and consistent identity preservation.}
\label{fig4}
\end{figure*}

\begin{table*}[!t]
\centering
\resizebox{\textwidth}{!}{
\begin{tabular}{lccccccccccccc}
\toprule
\multirow{2}{*}{\textbf{Model}} & \multicolumn{2}{c}{\textbf{Speed}} & \multicolumn{4}{c}{\textbf{Audio}} & \multicolumn{3}{c}{\textbf{Video}} & \multicolumn{2}{c}{\textbf{Audio-Video}} & \multicolumn{2}{c}{\textbf{Lip-Sync}} \\
\cmidrule(lr){2-3} \cmidrule(lr){4-7} \cmidrule(lr){8-10} \cmidrule(lr){11-12} \cmidrule(lr){13-14}
 & FPS\(\uparrow\) &TTFF\(\downarrow\) & AQ\(\uparrow\) & A-ID\(\uparrow\)& Speech\(\downarrow\) & TA\(\uparrow\) & VQ\(\uparrow\)& TV\(\uparrow\) & V-ID\(\uparrow\) & Sync-D\(\downarrow\) & IB-Score\(\uparrow\) & LSE-C\(\uparrow\) & LSE-D\(\downarrow\)\\
\midrule
Use CrossAttn & 26.17 & 3.53s & 5.85 & 0.504 & 0.115 & \textbf{0.501} & \textbf{3.95} & 25.03 & 0.954 & \textbf{0.038} & 0.336 & 3.84 & 8.81 \\
w/o DistillTeacher & 27.64 & 3.49s & \textbf{6.25} & 0.498 & 0.489 & 0.418 & 3.83 & 24.89 & 0.951 & 0.045 & 0.339 & 3.98 & 8.89 \\
w/o DataConstruct & 27.64 & 3.49s & 6.17 & 0.512 & 0.382 & 0.439 & \underline{3.93} & 25.12 & 0.956 & 0.051 & 0.347 & \textbf{4.38} & 8.94 \\
w/o GPC & 28.13 & 3.47s & 5.98 & 0.515 & 0.239 & 0.435 & 3.90 & 24.87 & \textbf{0.959} & 0.050 & 0.320 & 3.71 & 8.56 \\
\arrayrulecolor{gray!70}
\midrule
\arrayrulecolor{black}
w/o VisualIdentity & \underline{28.57} & \underline{3.43s} & 6.19 & \underline{0.520} & 0.121 & 0.441 & 3.81 & 25.02 & 0.949 & 0.050 & \textbf{0.372} & \underline{4.23} & \textbf{8.35} \\
w/o SpeakerIdentity & 27.64 & 3.49s & 6.05 & 0.478 & \underline{0.099} & 0.439 & 3.92 & \underline{25.16} & 0.953 & 0.045 & \underline{0.359} & 4.11 & 8.72 \\
w/o MRCM & \textbf{28.57} & \textbf{3.43s} & 6.13 & 0.462 & 0.104 & 0.442 & 3.87 & 25.08 & 0.949 & \underline{0.039} & 0.342 & 4.17 & \underline{8.54} \\
\arrayrulecolor{gray!70}
\midrule
\arrayrulecolor{black}
\textbf{OmniMate(Full)} & 27.64 & 3.49s & \underline{6.22} & \textbf{0.525} & \textbf{0.097} & \underline{0.449} & 3.93 & \textbf{25.19} & \underline{0.957} & 0.042 & 0.356 & 4.06 & 8.66 \\
\midrule
\textbf{OmniMate(30s)} & \textbf{27.64} & \textbf{3.49s} & 6.20 & 0.519 & 0.117 & \textbf{0.441} & \underline{3.95} & \underline{25.15} & \underline{0.955} & \textbf{0.038} & \textbf{0.370} & 4.03 & 8.69 \\
\textbf{OmniMate(60s)} & \textbf{27.64} & \textbf{3.49s} & \textbf{6.54} & \textbf{0.540} & \underline{0.103} & 0.402 & \textbf{3.97} & \textbf{25.18} & \textbf{0.957} & 0.046 & 0.351 & \textbf{4.10} & \underline{8.62} \\
\textbf{OmniMate(240s)} & \textbf{27.64} & \textbf{3.49s} & \underline{6.31} & \underline{0.527} & \textbf{0.099} & \underline{0.434} & 3.95 & 25.11 & 0.949 & \underline{0.041} & \underline{0.357} & \underline{4.07} & \textbf{8.57} \\
\bottomrule
\end{tabular}
}
\caption{Ablation studies of GPC and MRCM, including injection strategies, training methods, and component-wise analysis. We further evaluate performance across different generation durations.}
\label{tab:ablation}
\end{table*}

\begin{figure*}[t]
\centering
\includegraphics[width=2\columnwidth, trim=15 130 15 30, clip=true]{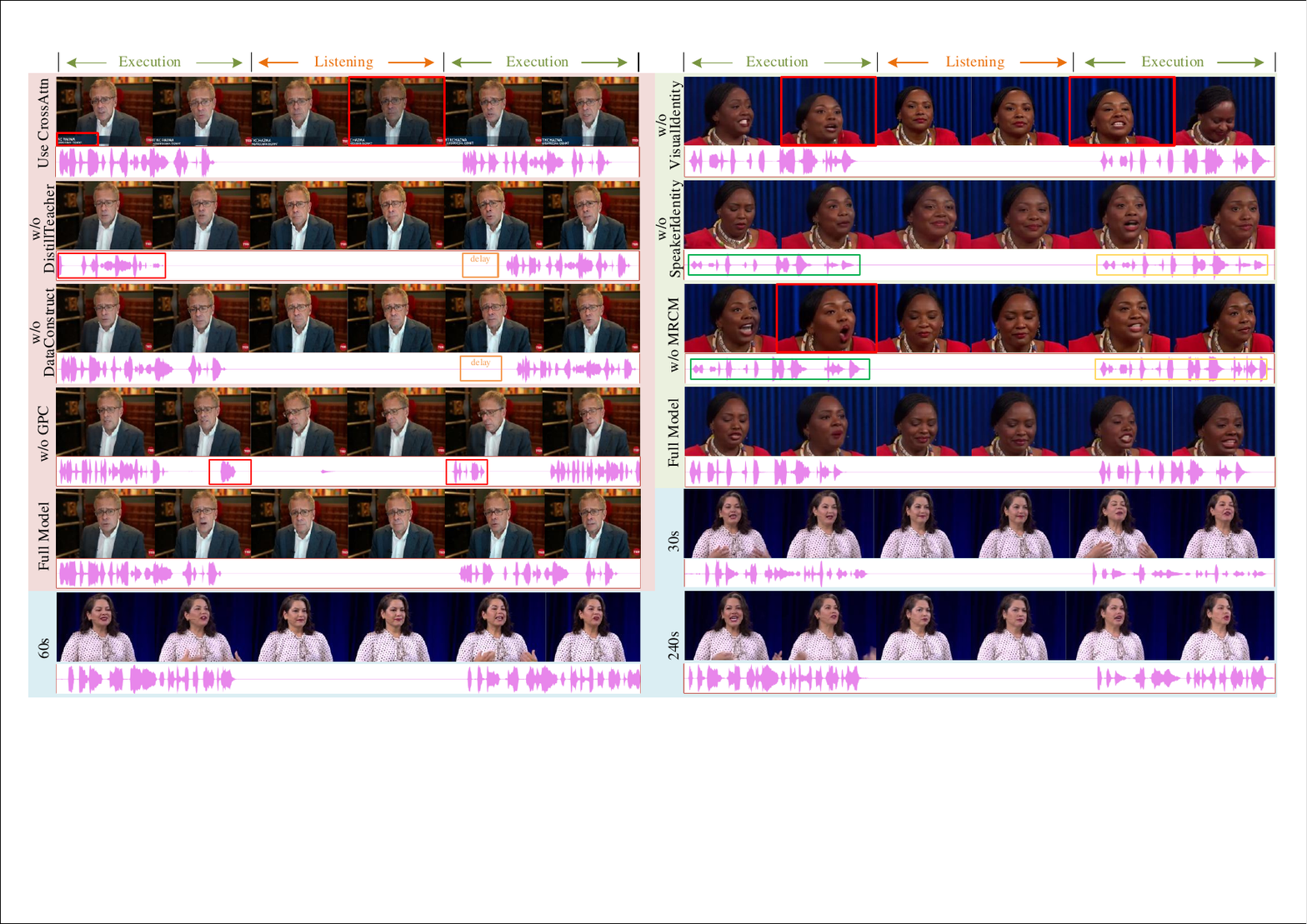} 
\caption{Qualitative ablation of OmniMate. Ablation studies demonstrate that our full model achieves more natural and accurate interactions, stronger visual and speaker identity consistency, and maintains stable quality during long-duration generation.}
\label{fig5}
\end{figure*}

\subsection{Experimental Setup}
Our framework is built upon LTX2.3. Training and inference are conducted with adaptive aspect ratios at 480p resolution, except for DMD distillation at 384p. In the streaming setting, the model generates 3 video latents and 25 audio latents per chunk. Training is performed on 32 NVIDIA H100 GPUs with four stages of 50K, 5K, 8K, and 5K steps, respectively. We employ FSDP for efficient training, with learning rates of $2\times10^{-5}$ for non-DMD stages and $2\times10^{-6}$/$4\times10^{-7}$ for the student/fake score branches during DMD. During inference, KV cache and pipeline parallelism are adopted to accelerate real-time streaming generation.

\subsection{Dataset}
We train our model on a large-scale audio-visual dataset containing approximately 3 million high-quality clips after filtering. The dataset consists of three sources. The first source includes public talking-head datasets, such as HDTF~\cite{zhang2021flow}, VFHQ~\cite{xie2022vfhq}, VoxCeleb2~\cite{chung2018voxceleb2}, CelebV-Text~\cite{yu2023celebv}, and AVSpeech~\cite{ephrat2018looking}, providing diverse identities and rich speech-driven facial motions. The second source OpenHumanVid~\cite{li2024openhumanvid} contains general video data from movies and TV shows, which improves the model's ability to handle diverse scenes and dynamics. The third source is our proprietary high-quality Chinese and English talking-head dataset, which further expands the data scale with diverse speakers, expressions, and long-duration conversational sequences for interactive streaming generation.

\subsection{Evaluation Data Construction and Metrics}
To evaluate our model's capabilities in interactive scenarios and its generative stability for long-duration videos, we construct a dedicated benchmark based on the character-scene subset of Versebench~\cite{wang2025universe1unifiedaudiovideogeneration}. Specifically, we augment the original prompts (treated as the execution state) with additional listening state prompts, and further design two state transitions to simulate realistic turn-based interactions between the avatar and user.

For evaluation metrics, we categorize them into five aspects: (1) generation speed, (2) audio and (3) video quality as well as its alignment with the input text, (4) audio-video synchronization, and (5) lip-sync accuracy. Further details are provided in the appendix.

\subsection{Comparison Results}
We compare OmniMate with existing audio-video generation models, including offline bidirectional-attention models Ovi-1.1~\cite{low2025ovi}, MOVA~\cite{team2026mova}, UniVerse-1~\cite{wang2025universe1unifiedaudiovideogeneration}, DaVinci-Base~\cite{chern2026speed} and LTX-2.3~\cite{hacohen2026ltx}, and real-time causal-attention models OmniForcing~\cite{su2026omniforcing} and Hallo-Live~\cite{li2026hallo}. 

For quantitative evaluation, as shown in Table~\ref{tab:result}, OmniMate achieves the best streaming efficiency, with the highest generation speed (FPS) and the lowest time-to-first-frame (TTFF), enabling real-time interactive generation. Meanwhile, it preserves strong audio and visual identity consistency (A-ID and V-ID), achieves the highest speech accuracy, and obtains the best cross-modal consistency (IB-Score) while maintaining competitive visual quality and lip-sync performance. These results demonstrate that OmniMate effectively balances generation quality and real-time responsiveness.

For qualitative evaluation, as seen in Figure~\ref{fig4}, existing methods exhibit clear limitations in long-form interactive generation. Ovi-1.1, DaVinci-Base, UniVerse-1, and Hallo-Live suffer from progressive visual degradation over time. Moreover, most baselines generate unnecessary sounds or motions during listening periods, and methods such as DaVinci-Base, OmniForcing, and Hallo-Live struggle with accurate transitions between listening and execution states. OmniForcing also shows inferior visual quality and lip-sync performance. In contrast, OmniMate enables seamless state transitions while maintaining high-fidelity visual generation, audio quality, and audio-visual synchronization throughout long-duration interactions, demonstrating its effectiveness for real-time interactive audio-visual generation.

\subsection{Ablation Studies}
We conduct ablation studies to evaluate the effectiveness of the proposed components and training strategies in OmniMate, including GPC and MRCM, quantitative and qualitative comparisons are presented in Table~\ref{tab:ablation} and Figure~\ref{fig5}, respectively. .

For GPC, we first replace the proposed latent addition-based progress injection with a text cross-attention based design (Use CrossAttn). This variant introduces visual artifacts, such as title-like overlays, and may incorrectly darken the screen during listening states. Using ODE trajectories generated by the original multi-step teacher instead of the DMD-distilled bidirectional teacher (w/o DistillTeacher) results in inaccurate transition boundaries, leading to unreliable initialization and incomplete response generation. Removing the state-transition data construction strategy (w/o Data Construct) prevents the model from effectively learning execution-listening transitions, causing delayed responses. Without GPC (w/o GPC), the model lacks explicit progress control, resulting in inaccurate state transitions, premature termination, and repetitive speech generation.

For MRCM, removing visual identity conditioning (w/o VisualIdentity) leads to degraded visual identity consistency under different viewpoints, while removing speaker identity conditioning (w/o SpeakerIdentity) results in increased variations in speaker characteristics across dialogue turns. Removing both components (w/o MRCM) further affects cross-modal identity consistency.

These results demonstrate that GPC and MRCM, together with their corresponding training strategies, are essential for stable interaction, long-term identity preservation, and high-quality streaming generation.

Furthermore, we evaluate the robustness of OmniMate under different generation durations (30s, 60s, and 240s). The results show no significant degradation in visual quality, audio quality, or cross-modal consistency as the generation length increases, demonstrating the capability of OmniMate for stable long-duration audio-visual generation.

\section{Conclusion}
In this paper, we presented OmniMate, a unified framework for open-ended real-time interactive audio-visual avatar generation. To address adaptive response termination and long-term cross-modal identity preservation in open-ended streaming, we introduced GPC and MRCM, enabling temporally aligned state transitions and consistent visual and speaker identities. Extensive experiments demonstrate that OmniMate achieves low-latency streaming generation with high audio-visual quality and robust identity consistency over long interactions.

\subsubsection{Limitations and Future Work}
Despite its effectiveness, OmniMate still has limitations. Extremely long responses or inaccurate duration estimation may occasionally cause word omissions or repetitions. Moreover, identity preservation currently limits the generation of actions involving large appearance or scene changes. Future work will focus on improving progress modeling, extending OmniMate toward autonomous long-horizon interactions, and exploring a unified understanding-generation architecture for smoother interactions.

\bibliography{aaai2027}


\end{document}